\documentclass{article}
\usepackage[preprint,nonatbib]{neurips_2023}

\usepackage{cite}
\usepackage{amsmath,amssymb,amsfonts}
\usepackage{algorithmic}
\usepackage{graphicx}
\usepackage{textcomp}
\usepackage[dvipsnames]{xcolor}

\usepackage{adjustbox}

\usepackage{tabularx}

\usepackage{amsmath}
\usepackage{amssymb}

\usepackage[T1]{fontenc}    
\usepackage{hyperref}
\usepackage[hyphenbreaks]{breakurl}
\usepackage{booktabs}       
\usepackage{nicefrac}       
\usepackage{microtype}      
\usepackage{multirow}

\usepackage{cleveref} 

\usepackage{float}

\usepackage{hyphenat}
\usepackage{xspace}

\usepackage{paralist}
\usepackage{enumitem}

\usepackage{amsthm}
\theoremstyle{plain}

\newcommand{\xhdr}[1]{\vspace{1.7mm}\noindent{{\bf #1.}}}

\newcommand{\ie}{{i.e.,}\xspace}
\newcommand{\eg}{{e.g.,}\xspace}

\newcommand{\Secref}[1]{Sec.~\ref{#1}}

\newcommand{\dashsecref}[2]{Sec.~\ref{#1}--\ref{#2}}

\usepackage{extdash}

\usepackage{bbm} 
\usepackage{bm}
\usepackage{mathtools}
\DeclarePairedDelimiter\abs{\lvert}{\rvert}

\usepackage{amsmath}

\DeclareMathOperator*{\argmax}{arg\,max}

\usepackage[normalem]{ulem} 

\usepackage{cryptocode}
\usepackage{wrapfig}

\usepackage{csvsimple}
\usepackage{booktabs,subcaption}

\usepackage{array}
\usepackage{ragged2e}
\newcolumntype{P}[1]{>{\RaggedRight\hspace{0pt}}p{#1}}
\newcolumntype{M}[1]{>{\RaggedRight\hspace{0pt}}m{#1}}
\newcolumntype{C}{>{\centering\arraybackslash}m{20mm}}

\usepackage{mathtools}
\DeclarePairedDelimiterX{\dbldiv}[2]{(}{)}{
  #1\;\delimsize\|\;#2%
}

\begin{document}

\title{SoK: Memorization in General-Purpose Large~Language~Models}

\makeatletter
\newcommand{\linebreakand}{%
  \end{@IEEEauthorhalign}
  \hfill\mbox{}\par
  \mbox{}\hfill\begin{@IEEEauthorhalign}
}
\makeatother

\author{%
  Valentin Hartmann \\
  EPFL \\
  \texttt{valentin.hartmann@epfl.ch} \\
  \And
  Anshuman Suri \\
  University of Virginia \\
  \texttt{anshuman@virginia.edu} \\
  \And
  Vincent Bindschaedler \\
  University of Florida \\
  \texttt{vbindsch@cise.ufl.edu} \\
  \And
  David Evans \\
  University of Virginia \\
  \texttt{evans@virginia.edu} \\
  \And
  Shruti Tople \\
  Microsoft Research \\
  \texttt{shruti.tople@microsoft.com} \\
  \And
  Robert West \\
  EPFL \\
  \texttt{robert.west@epfl.ch} \\
}


\maketitle

\begin{abstract}
Large Language Models (LLMs) are advancing at a remarkable pace, with myriad applications under development. Unlike most earlier machine learning models, they are no longer built for one specific application but are designed to excel in a wide range of tasks. A major part of this success is due to their huge training datasets and the unprecedented number of model parameters, which allow them to memorize large amounts of information contained in the training data.
This memorization goes beyond mere language, and encompasses information only present in a few documents. This is often desirable since it is necessary for performing tasks such as question answering, and therefore an important part of learning, but also brings a whole array of issues, from privacy and security to copyright and beyond.
LLMs can memorize short secrets in the training data, but can also memorize concepts like facts or writing styles that can be expressed in text in many different ways.
We propose a taxonomy for memorization in LLMs that covers verbatim text, facts, ideas and algorithms, writing styles, distributional properties, and alignment goals. We describe the implications of each type of memorization---both positive and negative---for model performance, privacy, security and confidentiality, copyright, and auditing, and ways to detect and prevent memorization.
We further highlight the challenges that arise from the predominant way of defining memorization with respect to model behavior instead of model weights, due to LLM-specific phenomena such as reasoning capabilities or differences between decoding algorithms.
Throughout the paper, we describe potential risks and opportunities arising from memorization in LLMs that we hope will motivate new research directions.
\end{abstract}


\section{Introduction} \label{sec:intro}

Large language models (LLMs)~such as ChatGPT and LLaMA have taken the world by storm, with a plethora of applications and widespread interest and use by the general public, researchers, and industry. Unlike previous machine learning (ML) models that were usually trained on comparatively small, curated datasets, large modern general-purpose generative models are trained on data that was not originally meant for model training, and is too large for careful manual curation.
Increasing model sizes brings an increased capability of models to memorize substantial parts of those training sets. Memorization, especially in LLMs, can have unintended consequences such as leakage of sensitive information like social security numbers \cite{lukas2023analyzing,kim2023propile}, or the regurgitation of large parts of training documents \cite{carlini2023quantifying,ozdayi-etal-2023-controlling}.

Research on the privacy and security of LLMs has so far mostly looked at such verbatim memorization, in line with prior work on membership inference~\cite{li2013membership, shokri2017membership} and model inversion/attribute inference~\cite{fredrikson2015model, wang2021improving}. We argue that text data contains information such as facts or writing styles that LLMs can memorize but are not captured by just considering memorized verbatim text. We aim to broaden the understanding around memorization in LLMs by providing a taxonomy of memorization. We further want to draw attention to risks and benefits of memorization for various domains.

\xhdr{Contributions} We systematize a wide range of works on the broad topic of memorization for large language models, discussing challenges with defining memorization and considerations that anyone aiming to measure memorization in a given setting has to take into account (\Secref{sec:overview}). We identify the following disparate types of information that LLMs can be said to memorize:
            \begin{inparaenum}[1)]
            \item verbatim text (\Secref{sec:verbatim}),
            \item facts (\Secref{sec:facts}),
            \item ideas and algorithms (\Secref{sec:ideas}),
            \item writing styles (\Secref{sec:styles}),
            \item properties of the training distribution (\Secref{sec:dist_prop}), and
            \item alignment goals (\Secref{sec:alignment}).
            \end{inparaenum}
    For each type of memorization, we provide possible definitions and ways to detect and prevent it. We further discuss the implications of these types of memorization for model performance, risks (privacy, security), and governance (copyright, auditing). 
    We summarize existing research, and highlight new research directions in the absence of prior work. Table~\ref{table:implications} overviews the most important implications.

    By bringing together research from LLMs, ML, privacy, security and law, we provide practitioners that want to train and deploy LLMs with an overview of the challenges associated with memorization. Researchers benefit from a new contextualization of existing research, and from the open problems and research ideas we identify within these contexts.

\renewcommand{\arraystretch}{1.5}
\begin{table*}[]
    \centering
    \begin{adjustbox}{center}
    \begin{tabular}{P{0.92in}|P{1in}P{0.85in}P{1.1in}P{1.1in}P{1.2in}}
    \toprule
    \multicolumn{1}{C}{Type of Memorization} & 
    \multicolumn{1}{C}{Performance, Alignment} & \multicolumn{1}{C}{Privacy} & \multicolumn{1}{C}{Security, \mbox{Confidentiality}} & \multicolumn{1}{C}{Copyright} & \multicolumn{1}{C}{Auditing}  \\
    \midrule
    Verbatim text (\Secref{sec:verbatim}) &
    capability to quote (+), ignoring instructions & leaking sensitive data & leaking proprietary information & regurgitating training data & watermarking (+)\\
    Facts (\Secref{sec:facts}) & answering questions (+), overriding prompt data & leaking personal information & leaking cryptographic keys & \multicolumn{1}{c}{$\varnothing$} & detecting hallucination (+)\\
    Ideas and algorithms (\Secref{sec:ideas}) & performing common algorithms (+), reproducing harmful ideas & de\-con\-tex\-tu\-a\-li\-za\-tion of irony & leaking secret ideas & violation through fan fiction & determining generalization capabilities (+)\\
    Writing styles (\Secref{sec:styles}) & style transfer (+), reproducing toxicity & author attribution & social engineering & violation in combination with general idea of a work & detecting underrepresented communities in training data (+)\\
    Properties of training distribution (\Secref{sec:dist_prop}) & necessary for learning (+) & identifying human labelers & theft of preprocessing parameters & help in identifying potential violations (+) & detecting biases (+)\\
    Alignment goals (\Secref{sec:alignment}) & necessary for alignment (+) & leaking preferences of human labelers & spotting weaknesses in safety alignment & \multicolumn{1}{c}{$\varnothing$} & verifying safety claims (+)\\
    \bottomrule
    \end{tabular}
    \end{adjustbox}
    \caption{Main potential implications of the kinds of memorization covered in this paper across different domains. Positive implications are marked with (+), negative implication are not marked. We use $\varnothing$ to demarcate domains where we did not identify any obvious risks or benefits.}
    \label{table:implications}
\end{table*}

\xhdr{Related work}
Memorization is often only considered explicitly in the context of verbatim memorization \cite{carlini2023quantifying}. Ishihara~\cite{ishihara2023training} gives an overview of methods for detecting and exploiting such memorization.
AlKhassimi et al.~\cite{alkhamissi2022review} provide a survey on LLMs as knowledge bases, \ie stores for accessing and editing facts.
Some papers \cite{bubeck2023sparks,josifoski2023flows} evaluate the ability of LLMs to solve competitive programming questions, which often requires the knowledge of particular efficient algorithms, but we are not aware of any work particularly looking at algorithm or idea memorization.
Tyo et al.~\cite{tyo2022state} describe and evaluate several methods for authorship attribution based on LMs, a task that requires (at least partially) memorizing writing styles.
Distribution inference, the field concerned with the memorization of distributional properties, is small and emerging, and thus there are no surveys on the topic relating to language models.
Wang et al.~\cite{wang2023aligning} describe existing methods for aligning LLMs and for evaluating alignment, which can be seen as the memorization of alignment goals.
Overall, there are only few works that explicitly consider memorization, and none that approach this topic with the breadth of our paper.

The domains on which we consider the impact of memorization, on the other hand, were picked precisely because they are of large interest.
Model performance is usually evaluated through extensive benchmarking, where BIG-bench \cite{srivastava2022beyond} particularly stands out due to its comprehensiveness.
Smith et al.~\cite{smith2023identifying} and Fan et al.~\cite{fan2023trustworthiness} survey different types of privacy risks emerging from LLMs.
Mozes et al.~\cite{mozes2023use} and Ferrara et al.~\cite{ferrara2023genai} discuss safety and other risks resulting from the malicious use of LLMs.
Henderson et al.~\cite{henderson2023foundation} outline copyright violation risks resulting form the deployment of foundation models such as pretrained LLMs.
Mökander et al.~\cite{mokander2023auditing} discuss and propose ways to audit LLMs, both in terms of process, and in terms of technical tools.
Despite the large body of research on those domains, our paper is the first to consider them through the lens of memorization.
\section{Background}
\label{sec:background}

Autoregressive models are trained to predict the next text token based on the sequence of previous tokens, as opposed to, \eg bidirectional masked language models such as BERT \cite{devlin2018bert}, which are conditioned on both left and right context.
In this work, we focus on large autoregressive language models (LLMs) with billions of parameters such as GPT-4 \cite{openai2023gpt4} or Llama 2 \cite{touvron2023llama2}. Sometimes we discuss work that uses models with fewer parameters or models that are not autoregressive. For these, we use the generic acronym LM. We write \(p(\cdot|x)\) for the probability distribution of an autoregressive model given a context \(x\).

\subsection{Training LLMs} 

\label{sec:llms_stages}
Regardless of the concrete model architecture, current LLM training typically encompasses three main stages: pretraining, supervised fine-tuning, and reinforcement learning from human feedback (RLHF).

\xhdr{Stage 1: Pretraining}
The training data in this stage consists of text documents. The LLM is trained to predict the next token of a document given a prefix from the document. The pretraining dataset usually consists of public data such as web pages crawled from the Internet, books, etc., though the data sources are not always disclosed \cite{openai2023gpt4}.

\xhdr{Stage 2: Supervised fine-tuning}
The second stage uses a dataset of prompts and responses. The model is trained in a supervised fashion to give the response corresponding to a prompt in this dataset.
The prompts and responses can come from public, task-specific datasets \cite{wei2021finetuned} or be written specifically for the model training. In the latter case, both the prompts and the responses can be written by human labelers, or the prompts can come from user requests to a language model \cite{ouyang2022training}, which allows for more closely modeling the prompt distribution upon deployment of the model.

\xhdr{Stage 3: Reinforcement learning from human feedback (RLHF)}
This stage consists of several steps. First, human labelers rate answers generated by the model in response to prompts. The ratings or labels can also be model-generated \cite{touvron2023llama2} or come from language model users \cite{ouyang2022training}. These ratings are then used to train a reward model, which is finally used for optimizing the LLM via reinforcement learning.
Sometimes automated tools are used in addition to human labelers \cite{openai2023gpt4}.

\subsection{Impact of LLM memorization}
\label{sec:implication_domains}

Memorization in LLMs impacts various domains. The implications of memorization for model performance and auditing are largely positive, whereas privacy, security and copyright are often negatively impacted. For each type of information in \dashsecref{sec:verbatim}{sec:alignment}, we describe the implications of memorization on the domains described below. The domains are grouped by task performance (model performance and alignment), risks (privacy; security and confidentiality), and governance (copyright; auditing).

\xhdr{Model performance and alignment} 
Model performance refers to the performance of the model in downstream applications, for example question answering, creative writing or code generation \cite{srivastava2022beyond}. This is usually not perfectly correlated with purely technical metrics like perplexity on a test set. The related problem of alignment refers to removing undesirable behavior of the model such as toxicity, bias, etc., and instilling desirable behavior such as instruction following \cite{bai2022training}.

\xhdr{Privacy}
These are typically risks associated with the release/inference of previously non-public information that can be linked to individuals. Even when models are trained only on public information, decontextualization of public information about individuals can be considered a privacy risk \cite{brown2022what}.

\xhdr{Security and confidentiality}
We consider potential harms through information leakage (e.g., leakage of API keys \cite{kulkarni2021github}) or other means for institutions that train or use LLMs, or contribute training data. This includes memorization that enables abuses of LLMs such as efficient social engineering attacks \cite{mozes2023use}.

\xhdr{Copyright}
\label{sec:copyright}
LLMs are typically trained on large datasets containing material with copyright protections, such as the BookCorpus dataset \cite{bandy2021addressing} (\eg BERT \cite{devlin2018bert} and GPT-3 \cite{brown2020language}), components like Books3 of The Pile \cite{gao2020pile} (\eg GPT-Neo \cite{black2021gpt} and Pythia \cite{biderman2023pythia}), or web crawls such as C4 \cite{raffel2020exploring} (\eg T5 \cite{raffel2020exploring} and LLaMA \cite{touvron2023llama1}). Typically, no licenses are obtained for the use of this material in model training.
In the US, the fair use doctrine (17 U.S.C. §107) allows for the use of copyrighted material without a license under certain conditions, but how this applies to training generative models is not yet clear.
Rahman and Santacana \cite{rahman2023beyond} argue that training an LLM on copyrighted documents itself does not constitute a copyright violation. They draw the analogy between a model learning from copyrighted texts and students learning from such texts. Lemley and Casey \cite{lemley2020fair} come to the same conclusion, arguing that the model weights constitute a transformation of the copyrighted training data. However, this debate is not completely settled yet \cite{henderson2023foundation}.
In applying the fair use defense to outputs produced by an LLM, Rahman and Santacana point to the importance of taking into account the context in which these outputs are used: educational purposes might fall under fair use, but commercial purposes not. 
Henderson et al.~\cite{henderson2023foundation} discuss the applicability of fair use to LLMs in depth, and likewise conclude that the fair use defense does not always apply.
Importantly, the authors give examples showing that verbatim copying is not necessary for a copyright violation. Direct translations, re-using fictional characters for fan fiction, or re-telling a story in a different setting may all be considered copyright violations.

\xhdr{Auditing}
Some forms of memorization can be used to audit LLMs, that is, to test their compliance with specifications or regulations pertaining to the composition of the training data or the model behavior \cite{mokander2023auditing}.
\section{Measuring memorization}
\label{sec:overview}

\label{subsec:memorization}
Learning from data requires extracting information from the data and generalizing from this information. The learning abilities required to perform the generalization vary for different tasks. Regurgitating a specific fact seen directly in the training data requires no generalization; mimicking the choice of words and syntactic structures that make up a person's writing style requires some generalization; writing a new creative story requires a lot of generalization. Our focus is on memorization, which we define roughly as learning that involves only little generalization.
We acknowledge that, as of now, there is no complete understanding of generalization \cite{zhang2021understanding} and no clear demarcation between memorization and generalization \cite{feldman2020learning,feldman2020neural}.
We do not attempt to make progress on this demarcation problem here, but rather just use our intuitive notion to scope our paper. The definitions that we consider in later sections are all concerned with more specific objects than general memorization.
In the remainder of this section, we describe general approaches for, and challenges with, identifying and measuring memorization.

\subsection{Inference attacks}
Privacy researchers often consider what information is stored in a model through inference attacks that are designed to make inferences about the training data. Two common types are \emph{membership inference} and \emph{attribute inference} \cite{yeom2018privacy}. In the membership inference task, an adversary is given access to a model and knowledge of a candidate record, and wants to determine whether or not that record was included in the model's training data. In attribute inference~\cite{wu2016methodology}, the adversary is given incomplete information about a record from the training set and has to use its access to the model to infer some missing information about this record. A successful membership inference attack implies that the target model has memorized \emph{something} about the candidate record, since the model reveals whether or not it was trained on that record, but not what precisely. 

A successful attribute inference attack, on the other hand, enables inferences about data not previously known to the adversary.\footnote{While a perfect membership inference attack can be used to perform attribute inference \cite{yeom2018privacy,salem2023sok} and thus implies that the model has memorized attributes of records, practical membership inference attacks are not perfect, and further usually only deployed for extracting one bit of information about membership or non-membership of a candidate record.}
This implies that the model has stored the information learned in some way. This does not necessarily mean that the model has \emph{memorized} these attributes, though. It could be that it has merely learned the data distribution well enough to predict the attributes based on other attributes of the record \cite{jayaraman2022attribute}. If, however, the attribute inference attack on a counterfactual model trained on the same data with the record removed is unsuccessful in recovering the attributes, then we have evidence that the attack's success was due to memorization of the target attributes by the original model. When successful and done carefully, this can give a much more precise characterization than membership inference of what the model has memorized about a record.
We consider memorization in the sense of both membership and attribute inference, since both types of memorization can have significant implications. For example, in the case of verbatim text, knowing that a particular text was used for training an LLM (a successful membership inference attack) might give an auditor information about whether a particular user's data was used in training.
Having a model output the first chapter of a book when prompted with its first sentence (a successful attribute inference attack) might constitute a copyright violation.

While membership and attribute inference are concerned with individual data points, we consider several pieces of information that may affect multiple documents, such as writing styles or parameters of preprocessing methods. Memorization of such information is better captured by the concept of \emph{distribution inference} (also called \emph{property inference}) \cite{ateniese2015hacking,suri2022formalizing,hartmann2023distribution}. Distribution inference assumes an adversary that has a set of candidate training distributions and aims to determine the underlying training distribution of a given model. For example, one might be interested in knowing whether a certain percentage of the data was written in a particular style, or what fraction of the software engineers described in the training data are female.

\subsection{Challenges in estimating memorization}
It is difficult to precisely determine whether a piece of information was memorized by an LLM. There are several reasons for this, which we discuss in this subsection.

\xhdr{Memorization vs.\ extractability}
Anything that a model knows about its training data needs to be stored in some way in the model's weights. A na\"{i}ve definition of memorization could thus be ``any information that is stored in the model's weights is memorized''.
However, evaluating this definition would be infeasible and basically amount to fully determining everything the model has learned.
Researchers thus resort to studying proxies for this memorization via extractability or discoverability, which only captures memorized information that can be accessed through known methods. This inherently underestimates memorization, since it assumes there are no better ways to extract information from the model. 

For example, many definitions are concerned with what can be inferred from model outputs---a subset of what can be inferred from model weights. This essentially assumes a black-box (API) adversary who has no direct access to the model. In some settings, such an assumption is valid and defenses such as output filtering, that do not prevent memorization in model weights but only the revelation of memorized information at prediction time, may be effective.
If output token probabilities are provided, definitions can make use of them (\eg the probabilities of the possible answers to a multiple choice question). Otherwise one has to resort to the model outputs in character space, which are produced by decoding algorithms. These decoding algorithms can be deterministic (\eg greedy decoding), or non-deterministic (\eg top-\(p\) sampling~\cite{holtzman2020curious}). Furthermore, the choice of decoding algorithm can influence model hallucination (see next paragraph) \cite{dziri2021neural,lee2022factuality} and other behaviors \cite{josifoski2022language}. Carlini et al.~\cite{carlini2023quantifying} find that swapping greedy decoding for beam search slightly increases the discoverability of memorized verbatim text. Another challenge with output-based definitions is their dependency on a prompt (with some exceptions \cite{carlini2021extracting}), the choice of which influences the amount of extracted memorized information \cite{jiang2020can}.

\xhdr{Hallucination}
LLMs can generate plausible content that cannot be inferred from their training or input data \cite{ji2023survey}, known as hallucination.
Causes of hallucination include training data that favors text generation that is not grounded in the data \cite{wang2020revisiting,dhingra2019handling}, or a training objective that differs from the task objective \cite{wang2020exposure}. Hallucination can also be linked to memorization \cite{raunak2021curious}. Hallucinations can make it seem as if the model had memorized a piece of information, even though it was not present in its training data. This is referred to as \emph{extrinsic hallucination}, as opposed to \emph{intrinsic hallucination} that contradicts the training data or input \cite{ji2023survey}.

\xhdr{Reasoning and generalization}
Beyond factual outputs that are not grounded in the training data, there are outputs that are grounded in the training data but not explicitly contained in it. For example, training documents might contain the facts ``\textsc{[a]} is a student of \textsc{[b]}'' and ``\textsc{[b]} is a professor at university \textsc{[x]}'', but not the fact ``\textsc{[a]} is a student at university \textsc{[x]}''. Still, if the LLM is sufficiently powerful to perform deductive reasoning, it will give the correct answer to the question ``Where does \textsc{[a]} study?'' \cite{huang2023towards}. Similarly, LLMs often appear to perform inductive reasoning \cite{yang2022language} such as guessing the nationality of a person based on their name \cite{poerner2020ebert} or generalizing from code seen during training to create novel algorithms \cite{josifoski2023flows}.

\xhdr{Distinguishing memorization from hallucination and reasoning}
A correct model response to a factual question does not reveal whether the model arrived at this response via generalization and reasoning, because of hallucination, or because it has memorized this response from the training data.
When looking for memorized information in the model weights instead of in the model behavior \cite{meng2022locating}, cases of reasoning and hallucination relating to input data \cite{huang2023towards,ji2023survey} can be avoided.
In many cases of interest, such as personal identifiers, social security numbers or long passages of verbatim text, it is unlikely that a model could hallucinate the target information or gain knowledge of it through reasoning. 

In the following five sections, we cover five different types of memorization, starting with verbatim text. Each of these sections is separated into an introduction, a subsection listing potential definitions, a subsection discussing the impact on various domains, a subsection on how to detect this type of memorization, and a final subsection that discusses mitigation strategies. 
\Secref{sec:mitigation} discusses potential mitigation strategies that are not specific to any type of memorization.
\section{Verbatim text}
\label{sec:verbatim}

Memorizing verbatim text is the most direct and low-level form of memorization. It is also quite prevalent---Carlini et al.~\cite{carlini2023quantifying} demonstrate how GPT-J~\cite{black2021gpt,wang2021gpt} memorized at least 1\% of its training data according to their extractability definition (\Secref{sec:verbatim_definitions}), which is similar to attribute inference.

There are different ways to look at verbatim memorization. Researchers have considered the memorization of entire training documents \cite{henderson2023foundation}, parts of training documents \cite{carlini2023quantifying} and, in the context of privacy risks from personally identifiable information (PII), the memorization of short sequences that comprise personal information such as names or email addresses \cite{lukas2023analyzing}. The concept of verbatim memorization can be broadened by also considering paraphrases, such as those resulting from the replacement of words with synonyms \cite{lee2023language}.
Note that the verbatim memorization of a text sequence that describes a fact or algorithm implies the memorization of this fact or algorithm, though in a low-level representation. We restrict this section mostly to aspects that are unique to verbatim memorization, and discuss memorizing facts and algorithms in later sections.

\subsection{Definitions}
\label{sec:verbatim_definitions}
Since verbatim memorization is related to the tasks of membership and attribute inference \cite{salem2023sok}, some definitions of inference attacks could also be applied to verbatim memorization. However, several formal definitions of specifically verbatim memorization in LLMs have been proposed, on which we focus here.

\xhdr{Exposure metric \cite{carlini2019secret}}
Carlini et al.\ explore memorization of out-of-distribution secrets by LMs. They consider strings of the form ``The random number is \(r\)'', where \(r\) is a number randomly sampled from some space \(\mathcal{R}\). The authors sample one particular \(r'\in\mathcal{R}\) and include the corresponding string in the training set of the LM. They define the \emph{exposure metric}, which essentially measures how many guesses an adversary that tries to guess \(r'\) saves by computing the perplexity of the string ``The random number is \(r\)'' for all \(r\in\mathcal{R}\) and guessing \(r\) in order of increasing perplexity, over guessing in a random order from \(\mathcal{R}\). This metric requires many queries to the model to compute, in addition to a retraining of the model. Note that this approach avoids the difficulty of determining what should and should not be learned by a model, since the artificial random strings are introduced as an explicit way to insert content that should not be learned.

\xhdr{Extractability \cite{carlini2021extracting}}
This definition by Carlini et al.\ defines a string \(y\) to be \emph{extractable} from an LM \(p\) if there exists a prefix \(x\) such that:
\(y\leftarrow \argmax_{y^{'}: \abs{y^{'}}=N} p(y~{'}\vert x)\).
Instead of the intractable computation of the \(\argmax\), the authors use a decoding algorithm such as greedy decoding in practice. They also point out that due to pathological cases any string could be extractable, \eg when prompting the model to repeat a given input string. However, they instantiate the definition only for cases where \(x\) is the start-of-sequence token or a prefix from a document.
The authors then specialize this definition to \emph{\(k\)-eidetic memorization}, which only allows for strings that are repeated at most \(k\) times in the training data.

\xhdr{Counterfactual memorization}
Following the observation that strings that occur more often in the training data are more likely to be reproduced by the model, Zhang et al.~\cite{zhang2021counterfactual} aim to disentangle the plurality of a document in the training data and the degree to which its verbatim text is memorized. To this end, they define counterfactual memorization for a document \(d\) as the expected difference in token prediction accuracy when predicting \(d\) with models trained on datasets containing \(d\) and datasets not containing \(d\). This definition is a variation on a definition by Feldman and Zhang for label memorization \cite{feldman2020neural}, and is closely related to the concept of algorithmic stability \cite{bousquet2002stability}. Counterfactual memorization does not measure memorization in any specific model, but the degree to which a given model architecture memorizes examples from a given distribution on average. It requires training multiple models, and also requires access to the training data. 

\subsection{Implications}
\xhdr{Model performance and alignment}
For certain types of tasks it is desirable that the model remembers verbatim text. E.g., reviews of a book can greatly benefit from verbatim quotes, and likewise news articles about political speeches.
On the other hand, verbatim memorization can also hurt model performance. As shown by McKenzie et al.~\cite{mckenzie2023inverse}, there are situations where LLMs ignore task instructions and output memorized text when the prompt contains text from the training data.

\xhdr{Privacy}
As discussed in \Secref{sec:llms_stages}, a model trainer may rely on user-submitted prompts and responses in training stages 2 and 3. These prompts can contain highly sensitive information, \eg when a user asks for advice on medical or relationship issues. In such cases, the prompts may contain detailed information about the user, allowing a third party to identify them in case the model regurgitates the prompt.

\xhdr{Security and confidentiality}
As with privacy, the security and confidentiality risks mostly come from the memorization of data from the training stages 2 and 3. LLMs used as coding assistants could leak code from non-public code bases of companies. This could give an advantage to competitors and make it easier for malicious actors to find vulnerabilities in applications based on this code. Leakage of sensitive information from code-based models has already been studied and established~\cite{niu2023codexleaks,huang2023not}. In fact, a recent incident with Samsung~\cite{mauran2023whoops} involved employees sharing internal, sensitive code and meeting notes with OpenAI's ChatGPT, where they may be used for training \cite{schade2023how}.
Similarly, LLMs integrated in office packages or explicitly prompted with content of internal documents of an organization could memorize and reveal secrets of the organization when trained on these documents.

\xhdr{Copyright}
As discussed in \Secref{sec:copyright}, it is not entirely clear yet whether verbatim memorization of copyrighted documents from training data in just the model weights itself constitutes a copyright infringement.
As further discussed there, even in case of regurgitation of copyrighted text, copyright violations can be context-dependent. At the same time, verbatim regurgitation may not always be necessary to constitute infringement. Lee et al.~\cite{lee2023language} investigate the related problem of plagiarism. They measure the degree of verbatim
(copying passages character by character),
paraphrase (paraphrasing sentences from a source document), and idea plagiarism (copying core ideas) performed by GPT-2 when prompting the model just with an end-of-sequence token.
The authors show that all three types of plagiarism, from both pretraining and fine-tuning data, occur both in pretrained and fine-tuned models both, but not for all fine-tuning datasets. Fine-tuning also seems to reliably eliminate verbatim plagiarism from pretraining data. Henderson et al.~\cite{henderson2023foundation} argue that preventing copyright violations can only be done on a higher semantic level that goes beyond verbatim text matching.

\xhdr{Auditing}
Detecting memorized verbatim text can enable identification of certain datasets used for the training of an LLM, although likely requiring access to the specific candidate dataset~\cite{carlini2023quantifying}. For example, the benchmark BIG-bench \cite{srivastava2022beyond} includes a randomly generated string that acts as a globally unique identifier (GUID), and a test that checks whether the model assigns an anomalously low/high probability to the GUID, which would indicate a contamination of the training data with data from the benchmark. Such canaries could be inserted by model trainers to detect theft or unlicensed use of their models, potentially aiding techniques like dataset inference~\cite{maini2020dataset} that might otherwise not work well with LLMs~\cite{szyller2023on}.
Memorized documents from different dates may also be used to determine a lower bound on the knowledge cutoff date (until when training data was collected) of a model, as recently demonstrated for Github's Copilot~\cite{debenedetti2023privacy}.

\subsection{Detecting memorized verbatim text}
\label{sec:verbatim_detecting}
There have been several attempts at detecting verbatim memorization that build upon membership inference tests. For instance, the current state-of-the-art membership inference attack for LLMs~\cite{mattern-etal-2023-membership} relies on fluctuations in loss values around neighborhood, which is similar in idea to Merlin~\cite{jayaraman2021revisiting}, a membership inference attack for general machine learning. However, it is important to keep in mind that successful membership inference does not necessarily imply verbatim memorization (see~\Secref{subsec:memorization}), or the converse.

While the cost to train LLMs makes it infeasible to utilize techniques that require training shadow models, as is done in most membership inference attacks, the open-endedness of prompts opens new avenues unavailable for other domains such as vision and tabular data. For instance, simply prompting the model with the title and author name of a training document~\cite{henderson2023foundation} is sometimes effective.
In the same work, Henderson et al.~\cite{henderson2023foundation} sample random snippets from books and show how some LLMs, when prompted with these snippets, return long sequences from the books. They show that instructions like \textit{replace every `a' with a `4' and `o' with a `0'} can circumvent content filters.
Yu et al.~\cite{pmlr-v202-yu23g} use prompts with function signatures and code comments to extract program code from LLMs.

Prompting techniques not explicitly designed for detecting memorization could be repurposed, such as adding prefixes to encourage grounding in the training data~\cite{weller2023according}. Instruction\hyp tuned models are much more amenable to this type of prompting, indicating that instruction tuning can make it easier to access information memorized in model weights.
Some attacks rely on prefix-suffix completion to detect memorization. Ozdayi et al.~\cite{ozdayi-etal-2023-controlling} use prompt tuning on models by utilizing white-box access and knowledge of some training records. The attack generates a prefix that can be prepended to a prompt to maximize the likelihood of generating a suffix corresponding to training data.
Outside of these prompting attacks, there have been some recent attempts at attributing memorization of examples to specific neurons in models~\cite{maini_can_2023}.

\subsection{Preventing memorization or extraction of verbatim text}
\label{sec:verbatim_preventing}

One strategy for preventing memorization of verbatim text is to avoid repetitions of verbatim text in the training data, motivated by the observation that the likelihood of a sequence is memorized increases with the number of times that sequence occurs in the training data. Carlini et al.~\cite{carlini2023quantifying} extract memorized verbatim text from models of the GPT-Neo \cite{black2021gpt,wang2021gpt} family. They sample text sequences from the training data, prompt the model with a prefix of each sequence and check whether the model generates the corresponding suffix. They find that the amount of memorized text increases with model size, repetition of the sequence in training data, and the length of the prefix prompt. Increased memorization from repeated sequences has been observed before, and consequently de\hyp duplication of the training data has been proposed as a countermeasure \cite{kandpal2022deduplicating,lee2021deduplicating}. However, this might run counter to the effective upsampling (via training for more epochs) of trustworthy sources commonly performed in the training of LLMs \cite{gao2020pile,touvron2023llama1}.

Mantri and Sasikumat \cite{mantri2023developing} propose several potential pathways towards LLMs that do not regurgitate memorized copyrighted content: pruning or zeroing out parameters associated with such content; fine-tuning the model with non-copyrighted content; and the use of loss functions that discourage the model from generating text too similar to copyrighted training data.
A more radical change to prevent verbatim memorization would be to use a substantially different training objective: instead of learning to predict individual tokens from training documents, the model could learn to predict the content of those documents at a higher semantic level. This has been tested for computer vision models, where the prediction of individual pixels in the training objective is replaced by latent representations of image patches~\cite{assran2023self}.

Verbatim memorization can also be addressed in black-box settings by using post-processing to block text sequences in the training data from occurring in the generated output.
Ippolito et al.~\cite{ippolito2022preventing} propose using a Bloom filter to detect n-grams of the model output that appear in training data, and re-generate tokens until there is no more match in the training data. However, the filter cannot account for small differences, such as changed whitespaces, and can be actively avoided by style-transfer prompts, \eg making the model respond in all lowercase.

\section{Facts} \label{sec:facts}

LLMs achieve good results on knowledge benchmarks even in closed book settings \cite{touvron2023llama1}, implying sufficient memorization of facts about the world. These can be facts about the real world like ``birds can fly'' or facts about fictional worlds like ``Harry Potter studies at Hogwarts'' \cite{chang2023speak}. Li et al.~\cite{li2023transformers} provide empirical evidence for memorization of the co-occurrence of words in different topical contexts in both embeddings and self-attention layers.

\subsection{Definitions}
\label{sec:facts_definitions}

Generic definitions of facts can capture the notion of what a fact is, but make it difficult to distinguish between desirable and undesirable learning of facts.

\xhdr{Tuple completion}
Meng et al.~\cite{meng2022locating} represent facts as tuples \(t=(s,r,o)\) of a subject \(s\), a relationship \(r\) and an object \(o\). They define memorization of a fact \((s,r,o)\) as the model completing the prompt `\(s\ r\)' with `\(o\)'.
The Knowledge Assessment Risk Ratio (KaRR)~\cite{dong2023statistical} considers the ratios between the probability of the correct object being generated by the model when given \(s\) and \(r\), and when given only \(s\) or only \(r\). This is aimed at removing the influence of the prior probability of the model for generating \(o\).

\xhdr{Counterfactual memorization}
The definition of counterfactual memorization \cite{zhang2021counterfactual} (see \Secref{sec:verbatim_definitions}) might also apply to facts: Instead of a specific document, one would remove all occurrences of a specific fact from the training data, and measure how this influences the knowledge of the model about the fact. This could help with determining whether a model knows a given fact because of memorization (in that case it would not know the fact anymore after the removal) or because of reasoning (in that case the model would still know the fact).

\xhdr{Personally Identifiable Information}
A particular class of facts that provide information about identifiable individuals is known as \emph{personally identifiable information} (PII).
Kim et al.~\cite{kim2023propile} draw a distinction between structured PII and unstructured PII.
Information in structured PII follows a fixed pattern, such as email addresses and phone numbers. Unstructured PII can be expressed in different ways.
\eg ``\textsc{[person 1]} is the parent of \textsc{[person 2]}'' contains the same information as ``\textsc{[person 2]} is \textsc{[person 1]}'s child''.

\xhdr{PII extractability}
Lukas et al.~\cite{lukas2023analyzing} define three variants of PII leakage across different threat models, which can be used to define PII memorization, but also the memorization of more general facts. The first definition is extractability, which is the probability of a piece of PII being contained in an output produced by an unprompted generation of the model.

\xhdr{PII reconstruction and inference}
The second and third definitions of Lukas et al.\ measure the model's ability to associate PII with a context. A sentence containing at least one piece of PII is chosen from the training data and the PII is replaced by a \textsc{[mask]} token, for example ``The police arrested \textsc{[mask]} near the White House on 8/20.'' In PII reconstruction, the (approximately) most likely PII replacement for \textsc{[mask]} under the model likelihood is compared with the PII in the original sentence.
PII inference differs from PII reconstruction in that the model only has to choose from a predefined set of candidates.

\xhdr{Linkability of PII}
Kim et al.~\cite{kim2023propile} argue that a random disclosure of some personal information without it being linked to an individual does not necessarily pose a privacy risk. For example, it might not be problematic if an LLM leaks an address without the name of the resident. They propose the definition of linkable PII leakage, which implies memorization of the connection between different pieces of personal information. The definition roughly states that if the likelihood of a piece of personal information \(a_1\) under the model increases when conditioning the model on other personal information \(a_2,\dots,a_n\) linked to the same data subject, the model links \(a_1\) to \(a_2,\dots,a_n\). This definition is also applicable to other facts, \eg linking a football player to her teammates.

\subsection{Implications}
\xhdr{Model performance and alignment}
In order to answer questions about the world, LLMs need to memorize facts about the world. Even when facts are retrieved from external sources \cite{pan2023unifying,ram2023context,shi2023replug,khandelwal2019generalization, nakano2021webgpt}, the LLM needs to have enough world knowledge to make sense of those facts. Likewise, there is information that is structurally indistinguishable from PII, but meant for public use and useful for the model to memorize. Examples for this are the phone numbers of emergency services \cite{kim2023propile} or the support email address of a company.
On the other hand, memorization of invalid logical theorems can lead to the model committing logical fallacies \cite{mckenzie2023inverse}. In some cases, it might not be desirable that the model uses facts from its training, but rather information provided in the prompt. Memorized facts can hinder the model in incorporating this new information, \eg when redefining mathematical constants~\cite{mckenzie2023inverse}.

\xhdr{Privacy}
If only publicly-accessible data is used for the training of LLMs, they cannot memorize facts that are not already publicly disclosed. However, as argued by Brown et al.~\cite{brown2022what}, LLMs have the potential to decontextualize information, that is, bring up the information in contexts which it was not intended for or without essential surrounding context.
LLMs can reveal PII and other information that is not meant to be public but found its way online---\eg through users who shared sensitive information in online forums via accounts that can be traced back to them.
PII could also make its way into the model's memory if user-submitted queries are used in stage 2 or 3 of the training. For example, a user might ask the model to draft a reply to a letter, where the letter might contain the user's name and address.

\xhdr{Security and confidentiality}
As with privacy, involuntarily published confidential information contained in the pretraining data can also pose problems for security and confidentiality. Leaking API keys and cryptographic secrets accidentally committed to GitHub, which is a popular source for pretraining data~\cite{gao2020pile,touvron2023llama1}, is a well-known problem~\cite{meli2019bad,kulkarni2021github}. This risk extends to secrets obtained and published by cybercriminals \cite{siboni2014cyberspace,gta2022}. While an accidentally committed private key might be quickly deleted from GitHub, it could stay in the model's memory indefinitely.
As for the memorization of verbatim text, confidential facts such as salaries and strategic decisions may also be memorized by the model if user-submitted data is used in stages 2 and 3 of the training process and an organization gives the LLM access to internal documents or code.

\xhdr{Copyright}
Facts cannot be protected by copyright, and only their expression can be protected \cite{henderson2023foundation}. For example, textbooks on the same topic often have a large overlap in terms of the facts they contain without posing any copyright issues.
On the other hand, preventing memorization of facts from a particular training document would make it impossible for the model to memorize the verbatim text of the document in its entirety, and might mitigate some copyright concerns. This could be relevant for works of fiction where one might want the model to learn the writing style, but not the details of the fictional world.

\xhdr{Auditing}
As with verbatim memorization, facts memorized by the model provide a lower bound on the knowledge cutoff date of the model: if the model has learned a fact about an event, it has seen documents at least up to the point when that event occurred.
If the downstream application requires truthfulness, it is important to detect untrue statements learned by the model \cite{lin2021truthful}. Work on attributing model outputs to training documents~\cite{park2023trak,weller2023according,min2023silo,rashkin2023measuring} can help in detecting hallucinations.

\subsection{Detecting memorized facts}
\label{sec:facts_detecting}
Commonly used benchmarks for assessing factual knowledge of LLMs consist of questions about facts in natural language \cite{kwiatkowski2019natural,joshi2017triviaqa}. Some are in the form of multiple-choice questions~\cite{hendrycks2020measuring}, which are mostly suited to determining aggregate knowledge of model. Another variant is \emph{cloze} tasks, where the model is asked to fill in masked-out entities~\cite{onishi2016did,petroni2019language,chang2023speak}. The exact prompt formulation can influence the success of knowledge extraction for cloze tasks \cite{jiang2020can} and questions \cite{srivastava2022beyond}. 

Jiang et al.~\cite{jiang2020can} propose ensemble methods that leverage multiple prompts, generated through mining and paraphrasing content from Wikipedia. Li et al.~\cite{li2023multi} demonstrate how jailbreaking models can be exploited to increase PII leakage, as opposed to standard querying. Dong et al.~\cite{dong2023statistical} treat subject, relationship and object in the tuple completion task (see \Secref{sec:facts_definitions}) as latent variables, and consider different textual aliases to generate different strings pertaining to the same \(s,r,o\) tuple. Jain et al.~\cite{jain2023bring} present the model with a fact and its negation, and compare the perplexity of the model on those two strings.
As a step towards identifying knowledge in the model's weights, Meng et al.~\cite{meng2022locating} use causal interventions to identify components of LLMs that store facts. Specifically, they consider \(s, r, o\) tuples, where the model has to predict the object from the subject and relationship.
Based on their findings, the authors posit that the MLP modules in the transformer act as two-layer key--value stores. Lehman et al.~\cite{lehman2021does} study medical PII leakage of name--condition pairs via the release of embedding weights.
Patil et al.~\cite{patil2023sensitive} demonstrate, via parameter-analysis and prompting-based methods, how PII leakage persists even after information ``deletion'' (based on parameter editing).

\subsection{Preventing memorization of facts}
\label{sec:facts_preventing}
Fact memorization may be mitigated at various stages in the training pipeline, including removing text associated with particular facts from the pretraining data through using machine unlearning techniques to remove specific learned facts from a model.

A technique called scrubbing acts already on the level of training documents---identifying pieces of text, \eg via named entity recognition~\cite{akbik2019flair,honnibal2020spacy}, and removing them or masking them with either a generic \textsc{[mask]} token or more specific tokens such as \textsc{[name]}.
Scrubbing has been used for removing PII \cite{lukas2023analyzing}, but could also be used for other types of facts (e.g., ``The capital of Germany is \textsc{[city]}.'').

Shi et al.~\cite{shi2022selective} propose a fine-grained variant of differential privacy (see ~\Secref{sec:mitigation_dp}) termed S-DP that works on the token level. Instead of protecting entire training documents with DP, it only protects tokens that have been identified as sensitive.
Their proposed method selectively adds privacy noise to gradients to which the protected tokens have contributed.

Removing memorized facts might be particularly important in some jurisdictions like the EU~\cite{eu2016gdpr} and California~\cite{california2018ccpa} that codify a right to be forgotten for individuals whose personal data has been used by a business.
Meng et al.~\cite{meng2022locating,meng2022mass} build on their hypothesis about key--values stores and propose a method to selectively change facts via values in those stores.

A general difficulty with preventing memorization of unstructured PII is that it requires a model with a deeper understanding of relationships within the training documents than for structured PII.
Eldan and Russinovich~\cite{eldan2023whos} demonstrate a technique for unlearning data from a particular source. The technique is based on a reference model fine-tuned on the source data, along with generic text auto-generated with GPT-4.
The KGA framework~\cite{wang2023kga} uses a process of unlearning to make the model’s performance on target data similar to unseen data, while maintaining overall performance. This approach has been successful with more abstract data sources, such as specific personas in chat-based datasets.
Bayazit et al.~\cite{bayazit2023discovering} develop a method to identify ``knowledge-critical'' subnetworks within GPT-2 models, using a joint loss function and demonstrate its effectiveness with concepts like ‘fruit’ and ‘swimming’.
\section{Ideas and algorithms} \label{sec:ideas}

Ideas are similar to facts in that there are multiple ways to express them in language. Ideas can be about the physical world such as reinforcing concrete with steel, as well as about fiction, such as a story around a boy whose parents got killed when he was small and who learns wizardry at a secret school.
Algorithms are a special type of idea that can be memorized in two ways: (1) as a description of a series of steps, similar to how the idea for the plot above can be memorized as a sequence of events; and (2) as a behavior, \ie the LLM implements the algorithm and executes it in its forward pass.
The former may occur via algorithm implementations in training data, the latter via input--output examples \cite{nanda2023progress}. We have not found any research on whether this also works the other way around. In this section, we only cover simple algorithms such as arithmetic operations that do not require a high level of generalization.
The distinction between facts and ideas can sometimes be difficult. `Harry Potter is a wizard [...]' is a fact of literature, and likewise the steps of photosynthesis a biological fact.

\subsection{Definitions}
\label{sec:ideas_definitions}
\xhdr{Tuple completion and linkability}
For certain types of ideas, a definition similar to tuple completion or linkability for facts (\Secref{sec:facts_definitions}) might be suitable. For ideas consisting of multiple elements (\eg multiple plot elements) or algorithms consisting of multiple steps, the model could be prompted with all but one element or step. Then one could check whether the true missing element or step is predicted by the model, and could define this as a memorization of the idea or algorithm. As with facts, though, care would have to be taken to ensure that a correct prediction is not due to generalization abilities of the model.

\xhdr{Input--output behavior}
Whether an LLM has memorized an algorithm in its behavior can be defined by whether it behaves correctly on all algorithm inputs \cite{nanda2023progress}. If the input space is too large, the requirement can be reduced to a correct output on a selected sample of inputs, as in the algorithm tasks\footnote{\url{https://github.com/google/BIG-bench/blob/main/bigbench/benchmark_tasks/keywords_to_tasks.md\#algorithms}} of the BIG-bench benchmark \cite{srivastava2022beyond}.

\subsection{Implications}
\xhdr{Model performance and alignment}
As with facts, one often wants the model to memorize ideas such as solutions to common problems or common tropes in fiction \cite{yuan2022wordcraft}. Similarly, one might want the model to be able to perform algorithms such as arithmetic manipulations \cite{saxton2019analysing}, or return an implementation of dynamic programming~\cite{bubeck2023sparks,josifoski2023flows}.
On the other hand, it is often undesirable if the model learns and reproduces harmful ideas. Meta considers three risk categories for its Llama~2 model \cite{touvron2023llama2}: illicit and criminal activities; hateful and harmful activities; and unqualified advice. Examples for ideas from all of these categories are very likely to be found in a dataset scraped from the Internet: plans for robberies in fictional stories; ideas around the superiority of one race  in online comments; or medical speculation by non-professionals in health forums.

\xhdr{Privacy}
Brown et al.~\cite{brown2022what} discuss the importance of the context in which information is shared. This can apply to ideas as well. For example, an individual might share an idea that is harmful or discriminatory on the Internet in an ironic way or as part of a fictional story. An LLM could, when asked about the stance of the person on that issue, attribute this idea to them without mentioning the context, opening them up to ostracism.

\xhdr{Security and confidentiality}
Companies hide new products from the public until they are officially presented. Partners in political coalitions often negotiate in private before presenting their compromise to the media. Using an LLM for research on technical questions around a new product or legal questions around political plans could lead to a leak of confidential ideas if the LLM's provider uses user-submitted queries for training, similar to how patentable ideas could be leaked.

\xhdr{Copyright}
Although abstract ideas typically do not enjoy copyright protection in the US \cite{elkin2023can}, there have been cases of fan fiction or use of fictional characters that constituted copyright violations~\cite{henderson2023foundation}.
Algorithms are not protected by copyright law \cite{bloch2022some,lemley2020fair}, but concrete implementations might be (see \Secref{sec:verbatim}).
Users might use an LLM for exploring or concretizing ideas that they consider patenting. If such user-submitted data is used in model training and the model memorizes these ideas, it might reveal them to other users. This could create prior art, making the granting of a patent impossible in many jurisdictions (EU: Article 54(3) EPC; US: 35 U.S.C. §102).

\xhdr{Auditing}
Determining how often an LLM generates new ideas instead of reproducing ideas from its training data would help determine the suitability of the LLM for tasks such as creative writing.
When using the algorithmic capabilities of an LLM, it is important to know whether it has just memorized many input--output examples, or has learned the underlying algorithm, which only happens after long enough training~\cite{nanda2023progress}.

\subsection{Detecting memorized ideas and algorithms}
As mentioned in \Secref{sec:ideas_definitions}, techniques for detecting memorized facts (\Secref{sec:facts_detecting}) might also apply to ideas.
The BIG-bench benchmark \cite{srivastava2022beyond} measures the capability of LLMs to perform common algorithms such as removing duplicates from a list of numbers or finding the longest common subsequence of two strings, by testing correctness of model responses on instances of these problems. Saxton et al.~\cite{saxton2019analysing} synthetically generate mathematics problems from fields such as algebra and arithmetic. They train models on smaller instances (\eg smaller numbers) and test them on larger instances to determine whether the models merely learn algorithms for the domain of the training examples, or learn the algorithms in their full generality. McCoy et al.~\cite{mccoy2023embers} compare the performance of LLMs on the same task, but with different parameters that do not change the task difficulty.

Stolfo et al.~\cite{stolfo2023understanding} perform causal mediation analysis on LLMs \cite{pearl2001direct} to get insights into how LLMs perform simple arithmetic operations. They change activation values to identify layers and neurons most responsible for those operations. In a similar manner, Wang et al.~\cite{wang2022interpretability} identify the circuit in GPT-2 responsible for detecting the grammatical object in a certain class of sentences. Nanda et al.~\cite{nanda2023progress} train a transformer to perform modular addition. Via careful inspection, they identify the exact algorithm that the model uses to perform this task. Interestingly, they find that in the beginning of training the model memorizes the training examples; in a second training phase it learns to perform the general algorithm; and in a third phase it removes the memorized components.

\subsection{Preventing memorization or extraction of ideas and algorithms}
Supervised fine-tuning, RLHF, and safety context distillation are often used \cite{touvron2023llama2} to deter a model from generating ideas from certain categories like criminal activities. The latter is a form of fine-tuning aimed to make the model behave as if prompts were preceded by a prompt instructing the model to, \eg only generate safe responses. While not explicitly designed to prevent the model from outputting \emph{memorized} ideas from these categories, this can be a side effect.
\section{Writing styles} \label{sec:styles}

LLMs like GPT-4 can write in rhymes or imitate the writing style of Shakespeare \cite{bubeck2023sparks}. Microsoft recently announced the ``Sound like me'' feature for their Copilot, allowing the LLM to write in the user's style when drafting emails \cite{spatarro2023announcing}. With writing style, we mean the linguistic definition of style~\cite{jin2022deep}---everything about a text that goes beyond pure semantics, including the choice of words and sentence structures, the characteristic use of stylistic devices such as alliterations and metaphors, and the level of formality.
We consider not only writing in natural languages, but also programming languages. There, with different styles we mean functionally equivalent pieces of code that differ in stylistic features like the naming of variables, their capitalization, the use of software design patterns, and formatting differences like the use of tabs or spaces for indentation.

\subsection{Definitions}
\xhdr{Mixture distribution}
Several authors \cite{andreas2022language,nardo2023waluigi} suggest that LMs learn to separate agents (\eg separated by different beliefs) in their training data. This concept is formalized by Wolf et al.~\cite{wolf2023fundamental}, who describe a language model as a mixture over different probability distributions.
We can instantiate this formalism for writing styles, where we describe the writing style of the model as a mixture of the writing styles of authors seen during training, one writing style per author (or, alternatively, one writing style per demographic group, \eg age groups or speakers of a dialect \cite[Sec.\ 4.1]{troiano2023theories}). The probability \(p(y)\) that the model assigns to a given string \(y\) can then be decomposed as \(p(y)=\sum_{\phi\in\Phi} w_{\phi}p_{\phi}(y)\), where each \(p_{\phi}\) corresponds to the writing style of one author in the training data and \(w_{\phi}\) is the prevalence of that style in the model outputs. We can define the memorization of one author's writing style as the presence of this writing style in the mixture, \ie that there is one \(p_{\phi}\) that corresponds to this author's writing style. Wolf et al. show that the model can be made to behave according to any \(p_{\phi}\) in the mixture by a suitably long prompt, given some technical conditions.
In practice, a prompt such as `Answer in the style of \textsc{[name]}' might suffice to isolate \(p_{\phi}\).

\xhdr{Authorship verification and attribution}
To determine whether a model can imitate an author's writing style sufficiently well, an \emph{authorship verification} (AV) or \emph{authorship attribution} (AA) task could be used \cite{tyo2022state}. In AV, an adversary is given two texts and has to determine whether they are written by the same author. In AA, an adversary is given texts from different authors and has to determine for a separate text by which of those authors it has been written. The advantage of an adversary in AV or AA over random guessing could be used a measure of the memorization of a writing style.

\subsection{Implications}
\xhdr{Model performance and alignment}
Reif et al.~\cite{reif2021recipe} show that LLMs are capable of augmented zero-shot style transfer, which can help in creative writing \cite{reif2021recipe} or tasks like using an LLM to make an email draft sound more formal \cite{lu2023bounding}. It is also common to directly ask a model to generate an output in a specific style \cite{lu2023bounding}.
On the other hand, models that memorize undesirable writing styles from their training data are prone to replicating them. Allen-Zhu and Li~\cite{allen2023physics} show that when pretraining a transformer with text that contains grammar mistakes, the model can respond to prompts that contain faulty grammar with text that contains grammar mistakes as well. A similar phenomenon might occur with LLMs used as coding assistants in a codebase with low-quality code.
The Common Crawl dataset\footnote{\url{https://commoncrawl.org/}}, a popular source for pretraining data, contains hate speech~\cite{luccioni2021what}, which models might imitate. Larger models are more likely to produce toxic content \cite{solaiman2021process}, maybe due to their larger ability to memorize toxic writing styles from the tails of the training distribution.

\xhdr{Privacy}
An LLM that has memorized the writing styles of individuals could be used for authorship attribution~\cite{mosteller1963inference,tyo2022state}. One might, for example, try to determine the author of an anonymous text, if documents written by that author under their name were part of the training data.
Authorship attribution via pretrained transformer models has been explored before \cite{fabien2020bertaa,barlas2020cross,tyo2022state}, although with smaller models like BERT that require fine-tuning on target authors' documents. LLMs that have seen documents from multiple authors already during pretraining might be usable off-the-shelf for authorship attribution, making this technique much more accessible.

Many LLMs will have seen instances of documents written by an author together with additional information about that author (columns by a journalist that contain biographical information about authors, etc.). They might hence be able to perform author profiling \cite{bevendorff2022overview,troiano2023theories}, i.e., given a document, determine attributes of the author such as age or gender. This has recently been demonstrated using Reddit comments \cite{staab2023beyond}.

\xhdr{Security and confidentiality}
If a model has memorized and is able to replicate an individual's writing style, it could be used for impersonation for social engineering attacks~\cite{mozes2023use}.
If the number of memorized writing styles is large due to the large number of documents seen during training, this process could be automated at scale \cite{darktrace2023generative,hazell2023large}.
For these types of attacks, it might not even be necessary to exactly copy the individual's writing style, but rather just the writing style within the individual's peer group~\cite{troiano2023theories}. Social engineering with voice cloning is already common enough for the FTC to issue an alert~\cite{puig2023scammers}, and could be made much worse with convincing style copying.

\xhdr{Copyright}
A writing style itself is not protected by copyright, as Henderson et al.~\cite{henderson2023foundation} state. However, they show with one example that a text in the writing style of an author whose general idea is close to that of a copyrighted work of that author may violate copyright. The model would have to memorize at least the general idea of a text in addition to the writing style for this type of copyright violation.

\xhdr{Auditing}
Assuming that, like verbatim text, less frequent writing styles are less prone to being memorized, the set of memorized writing styles would be an indicator of underrepresented communities in the training data. As an example, Dodge et al.~\cite{dodge2021documenting} found that in the popular C4 text corpus~\cite{raffel2020exploring}, African-American English and Hispanic\hyp aligned English are disproportionately affected by the blocklist used for data cleaning.

\subsection{Detecting memorized writing styles}
If a model can successfully apply the style of an author to its output, this is evidence for the memorization of the author's style. This could be measured by existing authorship verification or attribution methods \cite{tyo2022state}.
For more generic styles like `formal' or `poetic', researchers use zero-shot prompting \cite{lu2023bounding}, augmented zero-shot prompting (giving the model examples of other styles than the target one) or few-shot prompting \cite{reif2021recipe}.

\subsection{Preventing memorization or extraction of writing styles}
If a writing style's frequency affects memorization (as it is the case for Wikipedia entities~\cite{mallen2022not}), one solution is to limit the amount of text per author in the training data.
Soleiman et al.~\cite{solaiman2021process} demonstrate that fine-tuning on non-toxic human-written prompts can reduce toxicity, arguably a form of writing style. Likewise, toxicity can be reduced by RLHF \cite{ouyang2022training,wang2023decodingtrust} or incorporating human feedback in the pretraining objective \cite{korbak2023pretraining}.
Ilharco et al.~\cite{ilharco2023editing} fine-tune models specifically for undesirable behavior and subtract the weight difference from the original model to remove such behavior. Li et al.~\cite{li2023circuit} propose disconnecting model components to minimize loss on training data and maximize loss on unwanted behavior data. Both techniques could be applied to selectively remove writing styles. Li et al.'s technique could be applied to eliminate an individual author's writing style by dividing their documents into useful content and unimportant content, the latter serving as data for unwanted behavior.
Mireshghallah and Berg-Kirkpatrick~\cite{mireshghallah2021style} propose a VAE-based method for obfuscating the writing style of a text document by turning it into a generic style.
While the authors design this method to prevent discrimination and bias, it might be applied to the training corpus of an LLM to prevent the memorization of specific writing styles.
\section{Distributional properties of the training data} \label{sec:dist_prop}

\emph{Distribution inference}~\cite{suri2022formalizing,hartmann2023distribution}, also known as \emph{property inference}~\cite{ateniese2015hacking, ganju2018property}, is concerned with the leakage of (sensitive) properties of a model's training distribution. For LLMs, such properties of interest could be the proportion of documents authored by a given gender, the percentage of hateful content, the type of preprocessing used or data sources used in training.

\subsection{Definitions}
\xhdr{Distribution inference}
The degree to which a distributional property is memorized by a model architecture can be measured through a distribution inference game \cite{suri2022formalizing,hartmann2023distribution}, wherein an adversary is presented with a model that was trained on data sampled from a distribution with one specific value for the property, and has to guess that value. One then computes the advantage of the adversary over random guessing. This amount of leakage can also be quantified using the $n$-leaked metric~\cite{suri2023dissecting}, which captures how many records from the training distribution would need to be provided to leak a comparable amount of information about that distribution.

\subsection{Implications}

\xhdr{Model performance and alignment}
The entire premise of training is to obtain models that know something useful about the underlying training distribution, and generalize to similar yet unseen data. When the downstream task is well-defined, such as in regression or classification settings, one can---at least theoretically---make a clear distinction between properties of the training distribution that the model should learn and properties it can ignore~\cite{hartmann2023distribution}. However, general-purpose LLMs are not trained with a clearly demarcated set of downstream tasks in mind; the next-token-prediction objective is a proxy for a large, vaguely described set of tasks \cite{radford2018improving}. It is thus unclear which properties of the training distribution the model can safely ignore without impacting its utility.

\xhdr{Privacy}
Distributional membership inference could be used to identify individuals who contributed data in stages 2 and 3 of the model training \cite{hartmann2023distribution}. Distribution inference attacks for author inference have already been demonstrated for text classification models~\cite{melis2019exploiting}, and for fine-tuning data for LLMs~\cite{kandpal2023user}.
Distribution inference might further be used to inform attacks for detecting other kinds of memorization~\cite{zhou2022property}.

\xhdr{Security and confidentiality}
Distribution inference can be used to identify exact training data sources, enabling targeted poisoning attempts~\cite{carlini2023poisoning} via Sybil attacks.
It may further be possible to glean information about data preprocessing techniques.
In the ROOTS corpus~\cite{laurenccon2022bigscience} (the foundation for BLOOM~\cite{scao2022bloom}), for instance, code files are filtered out based on length and the percentage of alphanumeric characters. Finding good values for these hyperparameters requires compute or manual data inspection, making them worthwhile targets. Since these hyperparameters impact the training data distribution, a method for distribution inference might be able to reveal them.

\xhdr{Copyright}
When describing the training data distribution as a mixture over different data sources, one distributional property is whether a particular source---such as a specific website---is part of that mixture. This is described by distributional membership inference \cite{hartmann2023distribution}. Inferring that information from the model would allow an author to determine whether the model was potentially trained on their copyrighted documents.

\xhdr{Auditing}
The utility of distribution inference attacks, apart from inferring sensitive properties, also lies in auditing models without access to actual training data, which may not be available to the auditor. For instance, such attacks can be used to determine whether the training data is biased in any way, which is important, since tasks like hiring decisions and news generation require unbiased models. There has also been a demonstration of utilizing distribution inference (along with cryptographic primitives) for external auditing~\cite{duddu2023attesting} on classification models, focusing on attestation of gender and race-related properties.

\subsection{Detecting memorized distributional properties}
Most techniques for distribution inference rely on some form of shadow model training~\cite{ateniese2015hacking, ganju2018property, zhang2021leakage, suri2023dissecting}, which would be prohibitively expensive for LLMs. While there are a few techniques that do not require shadow models (like a simple loss-based estimation~\cite{suri2022formalizing}), the only settings in which such attacks demonstrate non-trivial leakage require data poisoning~\cite{chaudhari2023snap}. Additionally, all of these attacks make strong assumptions about the adversary's prior knowledge of the underlying training distribution, which is a general limitation of the framework. Distribution inference may be a realistic approach for learning about the data preprocessing, though. In a setting where the training data is known (\eg a public dataset such as The Pile~\cite{gao2020pile}), but not the preprocessing, an adversary can preprocess the same documents in different ways and track the loss of the target model on all variants. Intuitively, the loss should be lowest with the preprocessing used by the model trainer, even if the documents themselves were not part of the training data.

\subsection{Preventing memorization of distributional properties}
Learning techniques aimed at enhancing the privacy of individuals, such as differential privacy (\Secref{sec:mitigation_dp}), do not reduce distribution inference risk (and can, in fact, worsen it~\cite{suri2023dissecting}). Hartmann et al. \cite{hartmann2023distribution} propose general techniques for mitigating distribution inference attacks based on causal learning, specifically invariant risk minimization (IRM) \cite{arjovsky2019invariant}.
In a different context, techniques inspired by IRM have already been successfully used for making language models robust to different ways of preprocessing HTML pages in their training data~\cite{peyrard2022invariant}.
\section{Alignment goals} \label{sec:alignment}
To instill instruction-following, helpful, truthful and harmless behavior into the model, data generated by human labelers is used in training stages 2 and 3. Labelers write prompts and responses, and rate model outputs along various axes \cite{ouyang2022training,touvron2023llama2}, following guidelines provided by the model trainer. The effectiveness of the alignment training (see, \eg \cite{touvron2023llama2}), is proof that at least high-level goals from the guidelines such as harmlessness and helpfulness are memorized by the model.
Despite the guidelines, there is still disagreement between human labelers \cite{ouyang2022training,touvron2023llama2}.
Memorization from alignment training might thus not be limited to those guidelines---models might also memorize political, ethical and other opinions of labelers.

\subsection{Definitions}
\xhdr{Distribution inference}
The guidelines provided by the model trainer can be seen as a prior for the distribution of the data produced by the human labelers, \ie as a property of that distribution. One can ignore the pretraining by considering the training in stages 2 and 3 as training that is performed with a model initialized with the weights after stage 1. This leaves one with the standard distribution inference setting, where the distributional properties of interest are the alignment guidelines. Concretely, one could, for example, be interested in whether harmlessness is prioritized over helpfulness or whether the political guidelines to the labelers reflect a particular ideology.

\xhdr{Membership and attribute inference}
When one is interested in the text written or ratings given by individual labelers, membership and attribute inference are more apt frameworks. After all, the datasets consisting of prompt--completion pairs (stage 2) and of human ratings (stage 3) are simply datasets with each record contributed by one individual. As opposed to the standard setting, each labeler contributes multiple records to these datasets. Based on whether one is interested in one or all of the records from a labeler, one can use the item-level definition of membership inference or extensions to the user level \cite{levy2021learning}.

\subsection{Implications}

\xhdr{Model performance and alignment}
Supervised fine-tuning and RLHF or other methods based on human intervention \cite{rafailov2023direct,zhang2023wisdom,gulcehre2023reinforced} are currently essential for training aligned LLMs.
While part of the alignment training can be automated \cite{bai2022constitutional,openai2023gpt4,touvron2023llama2}, the alignment goals still need to be specified. It also seems unavoidable that in order to, \eg be able to refuse to follow certain requests, the model needs to memorize at least high-level alignment guidelines.
However, it seems desirable to have a model that \emph{only} memorizes the alignment goals but not the data of individual human labelers, in order to avoid any bias resulting from the make up of the group of labelers.

\xhdr{Privacy}
Preference ratings can reveal sensitive information about the labelers. For instance, truthfulness ratings of model outputs like ``Taiwan is a sovereign country'' might reveal political opinions. Similarly, the stance taken by a labeler in their response to the instruction ``Write an essay about whether US Americans should have the right to bear arms'' used for supervised fine-tuning. The often small number of labelers (\eg fewer than 20 for Llama~2~\cite{touvron2023llama2}) could make them easier targets for privacy attacks, since their individual contributions have a larger impact. It is also not uncommon to have their names mentioned in acknowledgements \cite{touvron2023llama2,ouyang2022training}.

\xhdr{Security and confidentiality}
Knowing the labeling guidelines or the reward function used for RLHF might help adversaries spot weaknesses in the safety alignment and eventually circumvent it. Stealing the reward function could also be interesting for competitors who want to train models and avoid the cost of human labelers.

\xhdr{Copyright}
Human labelers are likely so go through a formal process where they hand over the rights for their data to their employer. Copyright for this data, thus, may not be an issue. For instance, the participation agreement of Amazon Mechanical Turk~\cite{turk2023agreement} states that ``all ownership rights, including all intellectual property rights, will vest with that Requester''.

\xhdr{Auditing}
Analyzing the reward function can reveal model behaviors and properties the model trainer prioritizes. This should be reflected in the instructions given to the labelers, and subsequently in the reward function used for RLHF. Access to the reward function would thus enable an auditor to verify claims of a model trainer, for example that they prioritized harmlessness over helpfulness during model training.

\subsection{Detecting memorized alignment goals}
Alignment training happens in the later training stages, making it more prone to detection and extraction attacks than pretraining data \cite{feldman2018privacy,jagielski2023measuring}.
Since the training mode for supervised fine-tuning is similar to that for pretraining, attacks aimed at extracting verbatim text memorized during pretraining (\Secref{sec:verbatim_detecting}) might also work for the human-written responses to prompts used in stage 2. Different methods might be necessary for the prompts themselves, since the model is usually not directly trained to reproduce the prompts.

Regarding stage 3 training, it can be possible to reconstruct the reward function used for RLHF up to an additive, prompt-dependant constant when given access to the model after the stage 2 training, by using Eq.~5 of Rafailov et al.~\cite{rafailov2023direct}. If the model after stage 2 is not publicly available, but based on a publicly available pretrained model---this is, \eg the case for Llama~2 \cite{touvron2023llama2} and Mistral 7B~\cite{mistral2023mistral} --- one might try to approximate it through supervised fine-tuning by using one's own or public fine-tuning data such as the Flan Collection \cite{longpre2023flan}, used for LLaMA and Llama~2. Access to the reward function could be used to make inferences about labeler guidelines, for example, by checking whether harmful but helpful outputs are systematically higher ranked than harmless but non-helpful outputs. Individual labelers' data could be attacked using membership inference \cite{ye2022enhanced} or attribute inference \cite{jayaraman2022attribute} attacks.
Reuter and Schulze~\cite{reuter2023m} train a classifier to predict whether ChatGPT will refuse to answer a given prompt---behavior that is most likely predominantly learned in training stages 2 and 3.

\subsection{Preventing memorization or extraction of alignment goals}
The small number of human labelers might facilitate memorization of individual labelers' data, so increasing their number might reduce this risk.
Not releasing the model after stage 1 or 2 will make it harder for an adversary to determine which stage model behavior originates from---this does not prevent memorization (since the model is not changed in any way), but might help against specific attacks.

\section{General memorization mitigation strategies}
\label{sec:mitigation}

This section briefly discusses general-purpose strategies related to preventing memorization that are not specific to any particular type of memorization. Differential privacy (DP) (\Secref{sec:mitigation_dp}) and near access-freeness (NAF) (\Secref{sec:mitigation_naf}) are two frameworks originally designed to solve privacy and copyright problems, respectively, which aim at preventing certain forms of memorization (DP) or reproduction of memorized information (NAF). They have several shortcomings for preventing different types of memorization in LLMs though, as we discuss next. Elkien-Koren et al.~\cite{elkin2023can} argue in a very similar way why the frameworks are not suitable for preventing copyright violations. \Secref{sec:copyright_mitigation} discusses strategies that aim to mitigate copyright infringement with techniques that are external to the model.

\subsection{Differential privacy}
\label{sec:mitigation_dp}
Differential privacy (DP) \cite{dwork2006calibrating} is a privacy definition that can be satisfied by mechanisms where noise randomly sampled from appropriate distributions is incorporate in the training process to bound the impact of any individual record. Because of the random noise, the trained parameters will be nearly indistinguishable whether or not any one record was part of the training data. This makes it impossible for the model to remember any one training record. DP thus prevents the memorization of verbatim text, at least as long as this text is only contained in one training document. 

For other types of memorization, DP mechanisms can, however, be both underexhaustive and overexhaustive. Underexhaustive because facts or writing styles may be present in many different training documents, books can be contained indirectly in the training data through quotes, reviews, etc. \cite{chang2023speak}; and overexhaustive because even if one only wants to prevent the model from memorizing specific pieces of information such as PII or facts, DP adds noise to all information contained in a training document.

While underexhaustiveness could be addressed by adjusting the unit of privacy so DP mechanisms will provide privacy with respect to multiple records \cite{dwork2006differential}, this would still require identifying the number of documents that contain a particular item (which might be expressed in different ways), and might significantly worsen the DP-induced performance drop due to larger amounts of added noise. Even for DP at the level of individual documents, El-Mhamdi et al.~\cite{elmhamdi2022impossibility} argue that high-dimensional differentially private learning on heterogeneous data such as Internet text is impossible with high accuracy due to mean estimation being impossible under these conditions. The method by Shi et al.~\cite{shi2022selective} (\Secref{sec:facts_preventing}) that applies DP more selectively at a token level might help with the problem of overexhaustiveness, but requires exactly identifying the relevant tokens (\eg those that contain a fact), but is not easily applicable to memorization which is not concentrated in a few tokens of a document (\eg a writing style).

\subsection{Near access-freeness} \label{sec:mitigation_naf}
Vyas et al.~\cite{vyas2023provable} propose the notion of near access-free (NAF) generative models, aimed at preventing copyright violations. Given an subset \(\mathcal{C}\) of its training documents, a model \(p\) is NAF with respect to \(\mathcal{C}\) if for every \(C\in\mathcal{C}\) the difference between the model's output distribution and the output distribution of a specific model that was not trained on \(C\) is bounded by a fixed constant. If \(C\) only occurs once in the training data, this ensures that \(p\) is unlikely to output substantial parts of \(C\), unless those parts could have also been produced without access to \(C\)---which would not constitute a copyright violation. Vyas et al.\ also provide an extension for multiple occurrences.

Beyond verbatim text, NAF with the right distance metric and sufficiently small bound on the distance could also prevent the model from outputting PII such as social security numbers, or facts, as long as one knows how often they are contained in the training data. However, NAF has the same problems of under- and overexhaustiveness as DP. It is challenging to determine in how many documents a piece of information is contained, and prone to removing benign information that is unique to one or a few documents but that it is desirable for the model to learn. For the latter, consider the case where \(C\) is a book. Then a model without access to \(C\) would have a very low probability of correctly answering questions about the plot of \(C\), so an NAF model would not be able to answer those questions either.
Note that, unlike DP which is typically applied to the training process, NAF restricts the output distribution of the model. It can therefore be used to give guarantees for memorized information in the model outputs, but not for the memorization itself, \ie someone with access to the model weights might still be able to detect memorized information.
\subsection{Strategies for mitigating copyright violations via infrastructure}
\label{sec:copyright_mitigation}

In some cases, copyright violations may be prevented by preventing particular types of memorization, as discussed in the previous sections. There have also been some proposals for mitigating copyright violations through appropriate infrastructure. 

The most straightforward way to avoid copyright violations is by training only on data with permissive licenses \cite{kocetkov2022stack,fried2023incoder}. However, excluding data with non-permissive or unspecified licenses could significantly reduce the amount of available data \cite{min2023silo} and might exclude high-quality data sources such as textbooks.

Min et al.~\cite{min2023silo} propose training a LLM only on permissively licensed documents, and augmenting it with a datastore of copyrighted documents. This datastore can then be used at prediction time to improve LLM performance on domains not covered by the training data. This setup allows for precisely pinpointing which copyrighted documents contributed to a particular model output, and for easily removing copyrighted documents if required. Determining copyrighted documents that were used in producing a model output via this and other methods \cite{koh2017understanding} could allow for using documents with licenses that require author attribution \cite{henderson2023foundation}.
Ippolito and Yu~\cite{ippolito2023donottrain} describe a protocol similar to \texttt{robots.txt} wherein website owners could signal to model trainers via a file in the root directory which parts of their website are appropriate for model training, a variant of which has already been implemented by OpenAI \cite{gptbot} and Google \cite{googlebot}. In addition to deploying block requests \cite{newsblockers2023}, the New York Times has updated its terms of service to explicitly ban the use of its data for training ML models \cite{nytimesban2023}.

\section{Conclusions and open research questions}
In this paper, we have provided a taxonomy of memorization that goes beyond verbatim text. We have discussed challenges with defining memorization, and the consequences of memorization for various domains.
Throughout the paper, we have identified several gaps in the research literature, and proposed ideas for future work. We highlight three research directions that we deem particularly important:
\begin{itemize}
    \item Many existing definitions of memorization are specific to known attacks, and not the other way around. For example, the definitions that involve prompting the model with a prefix are adapted to the auto-regressive training of models, and, while useful for measuring a certain notion of memorization, do not cover adversaries that know a suffix of a document and want to extract a prefix.
    \item We did not find any research looking at memorization of data from the supervised fine-tuning and RLHF phase, even though these phases seem to contain the most severe memorization risks. In particular, it would be important to know whether user-submitted prompts that are used in these training phases can be extracted from the model.
    \item We want to minimize the amount of undesirable information memorized by LLMs due to the risks outlined in this paper. However, it is not clear which information LLMs need to memorize for good performance on downstream tasks, and the distinction between desirable and harmful learning is not well defined. This becomes an even more interesting question in light of the advent of LLMs enhanced with external knowledge, such as LLMs with Internet access.
\end{itemize}

Memorization in LLMs is an emerging area, and interactions between technical advances, demonstrated harms, and regulations and lawsuits will be necessary to achieve clarity on many of the issues raised in this paper. We expect this will be a long and evolving process, but one in which the research community should play an essential role.

\bibliographystyle{abbrv}
\bibliography{references}

\end{document}